\documentclass[10pt,conference,onecolumn]{IEEEtran}

\usepackage[pdftex]{graphicx}
\usepackage[justification=centering]{caption}
\usepackage{cite}
\usepackage{booktabs}
\usepackage{amsmath}
\usepackage{algorithm}
\usepackage[noend]{algpseudocode}
\usepackage[
singlelinecheck=false 
]{caption}

\makeatletter
\def\BState{\State\hskip-\ALG@thistlm}
\makeatother

\usepackage{xcolor}

\newcommand{\subparagraph}{}
\usepackage{titlesec}

\titleformat{\paragraph}
{\normalfont\normalsize\bfseries}{\theparagraph}{1em}{}
\titlespacing*{\paragraph}
{0pt}{3.25ex plus 1ex minus .2ex}{1.5ex plus .2ex}

\begin{document}

\title{Analytical Inverse Kinematic Solution for "Moz1" NonSRS 7-DOF Robot arm with novel arm angle}

\author{Ke CHEN}

{}
\IEEEtitleabstractindextext{%

\begin{abstract}
This paper presents an analytical solution to the inverse kinematic problem(IKP) for the seven degree-of-freedom (7-DOF)
Moz1 Robot Arm with offsets on wrist. We provide closed-form solutions with the novel arm angle
. it allow fully self-motion and solve the problem of algorithmic singularities within the
workspace.  It also provides information on how the redundancy is resolved in a new arm angle representation where traditional SEW angle faied to be defined 
and how singularities are handled. The solution is simple, fast
and exact, providing full solution space (i.e. all 16 solutions)
per pose.
\end{abstract}

\begin{IEEEkeywords}
7-DOF redundant robot arm, Closed-form Inverse kinematics, NonSRS structure, novel redundancy representation
\end{IEEEkeywords}}

\maketitle
\IEEEdisplaynontitleabstractindextext
\IEEEpeerreviewmaketitle
\section{Introduction}
\label{sec:introduction}

\IEEEPARstart{R}{esearch} on light-weight redundant manipulators, has grown in various
directions like human robot interaction \cite{kirschner2021involuntary} or machine learning \cite{petrik2021learning}. While some of these applications use the inverse
kinematics solver which is integrated in the robot’s control
system, others use a numerical solver. The former solution is
only available in position control mode and the code of the
robot control unit is closed source and information on the used algorithms and
therefore the redundancy resolution can hardly be found.
Numerical solvers, on the other hand, always come with a
risk of violating real-time constraints or not finding a solution
at all, although solutions exist in the workspace\cite{beeson2015trac}.\\
Finding an analytical solution for the IKP of serial robotic
manipulators is one of the fundamental basic problems in
robotics and there is still no general solution available.
There exist many analytical solutions for specific classes of
robots like 7R anthropomorphic robot arms with spherical
shoulder, spherical wrist(SRS structure), and three intersecting joint axes in
the elbow \cite{liu2017analytical}, \cite{bongardt2017inverse}. Although NonSRS structure manipulation(like Franka Emika Panda) is also
a 7R anthropomorphic robot arm, the previously mentioned
solutions do not apply for the robot since there are linear
offsets in its elbow and wrist joints. These offsets make
the analytical solution more complex compared to other
anthropomorphic arms like the KUKA iiwa, where all joints
are aligned in the zero configuration. More details on the
robot’s structure are given in section II.\\
In case of the 7dof robot, for SRS strucure, the IKP is solved based on the SEW angle as an redundancy represent\cite{shimizu2008analytical},
The IKP of some non-zero link offsets robot(NonSRS sturcture) could also be analytically solved based on the SEW angle method\cite{an2014analytical}, it required that the shoulder and wrist must be static during self-motion(we call it static-SW 7dof robot) 
However, for those nonstatic-SW robot, To the best of the author’s knowledge, there is no analytical IKP implementation available. \\
This paper provides both the analytical solution as well as a
real-time safe implementation for the nonstatic-SW 7dof robot. More details about inverse kinematic are given in section III, first in a specific configuration then promote it to general case.  \\
In addition, the singularity of this robot is analyzed and classified with reciprocity-based method, details on the singularity are presented
in section IV.
A conclusion of this work is given in section V, where
also some other robots with open IKP are mentioned and
possible future work based on this solution is illustrated.

\begin{figure}[ht]
      \centering
      \includegraphics[scale=0.4]{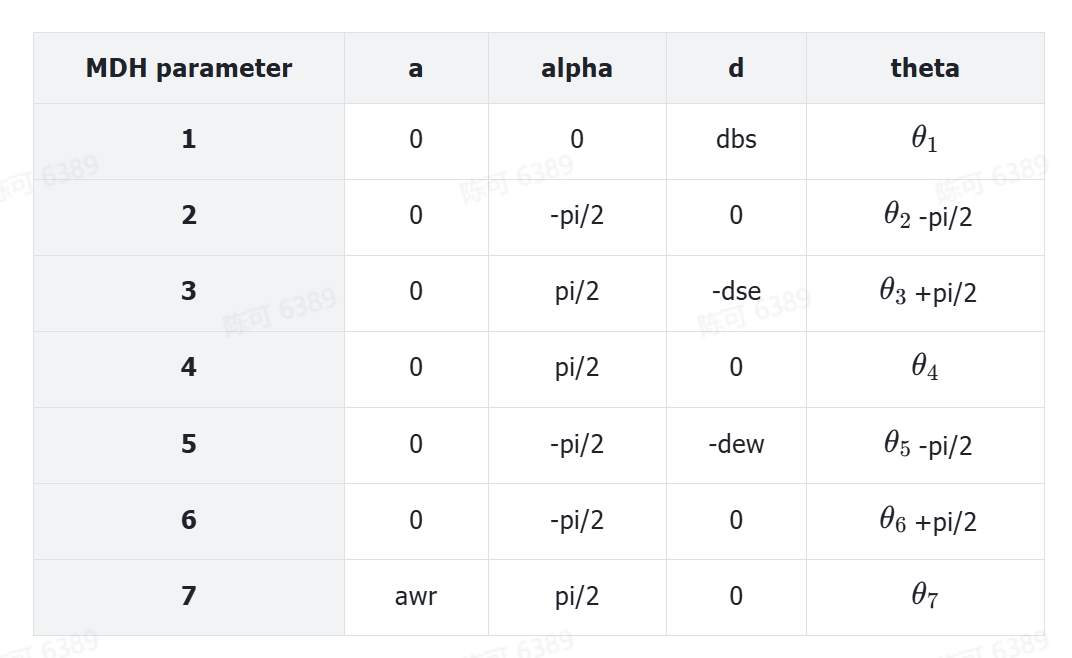}
      \caption{Robot MDH parameters}
      \label{fig:robot_localization_costmap}
\end{figure}

\begin{figure}[ht]
      \centering
      \includegraphics[scale=0.4]{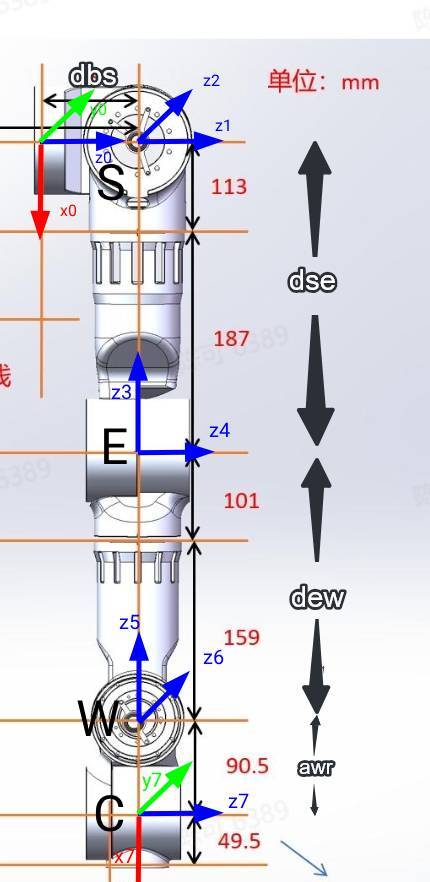}
      \caption{Robot kinematic structure}
      \label{fig1:robot_localization_costmap}
\end{figure}

\begin{figure}[ht]
      \centering
      \includegraphics[scale=0.2]{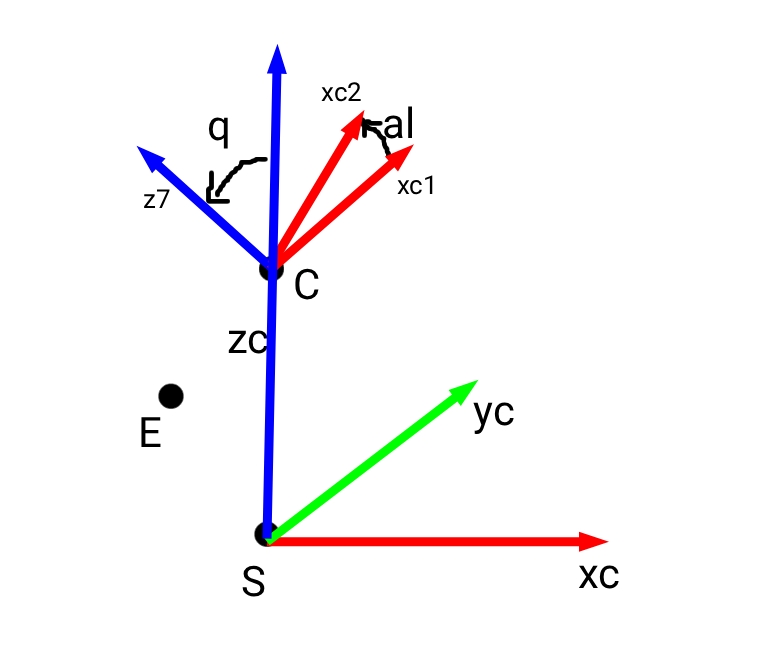}
      \caption{specific configuration}
      \label{fig2:robot_localization_costmap}
\end{figure}

\section{KINEMATIC STRUCTURE}
The Moz1 robot consists of eight links connected
by seven joints, which make the serial robot a redundant
manipulator. figure 1 show the MDH parameters
and figure 2 illustrates its kinematic structure. 
The shoulder is constructed similar to other anthropomorphic arms like the KUKA iiwa, i.e., it is also a spherical
shoulder. 
the first six joint is Similar to KUKA iiwa while there is an offset bettwen
sixth joint and seventh joint, so their axis are not intersecting.
which can be seen in figure 2. the offset
awr = 0.0905 m . 

The  fig. 2 shows the seven joint frames ${F}_{i}$
 for $0 \leq i \leq 6$ and j = i + 1 as well as the base frame $F_{0}$. A joint frame is represented by a
homogeneous transformation matrix as
\begin{equation}
      ^{i}T_{j} =  \begin{pmatrix}
^{i}R_{j} & ^{i}P_{j} \\
0 & 1 \\
\end{pmatrix}
\end{equation}
it describes the transformation between the joint connecting links
i and j. 
Additionally, some special frames are defined which are
important for later calculations. For analogies with anthropomorphic arms, 
a shoulder frame $F_{S}$ = $^{0}T_{2}$, an elbow
frame $F_{E}$ = $^{0}T_{4}$, and a wrist frame $F_{W}$ = $^{0}T_{6}$ are defined.

\section{ INVERSE KINEMATICS}
\subsection{Definition of the novel arm angle}
From Figure 2, there are four important points shoulder center S, elbow center E, wrist center W and center point of 7th axis C. Obvously, in the self motion of the robot arm, the wrist point W would move in a circle around point C, which is not suitable using SEW angle where the S and W is static
during the self motion. here we provide a new arm anlge using point S, E and C as SC are staic during the self motion. 
the reference plane is spaned by the vector SC and  $\overrightarrow{z_{7}}$ while the arm plane is spaned by the vector SC and vector SE, hence the arm angle would be the angle bettwen reference plane and arm plane around SC
$$ \psi = \angle plane<SC, \overrightarrow{z_{7}}>  \ plane<SC, SE> $$ 
\subsection{Special configuration}
First, let's consider a special configuration where the wrist center point C is located on
the base z-axis straight above the shoulder S, and the base x-axis is in the reference plane. As the figure 3 shows,  the angle from SC to  $\overrightarrow{z_{7}}$ around base y-axis is 
denoted as q, dsc is the length of vector SC, xc2 =$ y_{0}\times z_{7}$  and al is the angle from xc2 to xc1 =$x_{7}$ around $z_{7}$. In such configuration, we could present endframe in only 3 dof, the parameter dsc, q, and al as follow:
\begin{equation}
\begin{aligned}
^{0}T_{7}&=Trans(0,0,dbs+dsc)*Roty(q)*Rotz(al)\\
         &=\begin{bmatrix}
cos(al)*cos(q) & -cos(q)*sin(al)  & sin(q) & 0 \\
sin(al) & cos(al) & 0 & 0 \\
-cos(al)*sin(q) & sin(al)*sin(q) &cos(q)  &dbs+dsc  \\
0 & 0 & 0 & 1 \\
           \end{bmatrix} 
\end{aligned}
\end{equation}
given the endframe $^0T_{7}(dsc,q,al)$ and arm angle $\psi$, we present the analytical IK by algebraic method.

\subsubsection{solve q6 and q7}
\paragraph{arm equation}
denote the point E in base frame $^0E=(E_{x},E_{y},E_{z},1)$ According to the figure 3, we get the arm angle constrains 
\begin{equation}
\psi = atan2(E_{y},E_{x})+\pi
\label{e1}
\end{equation}
E in Frame $F_{7}$ is $$ ^{7}E = ^{7}T_{6}*^{6}T_{5}*^{5}T_{4}*\begin{bmatrix}
0 \\
0 \\
0 \\
1
\end{bmatrix}=\begin{bmatrix}
-t_{6} cos(q7) \\
t_{6} sin(q7) \\
-r_{6} \\
1
\end{bmatrix} $$ where $t_{6} = a_{wr}+d_{ew}cos(q6) $ and $r_{6} = d_{ew}sin(q6)$ \\
hence \begin{equation}
^{0}E =\  ^{0}T_{7}\ ^{7}E = \begin{bmatrix}
- r_{6}*sin(q) -t_{6}cos(q)cos(q7 - al) \\
 t_{6}sin(q7-al) \\
 dbs + dsc - r_{6}cos(q) + t_{6}sin(q)cos(q7-al) \\
1
\end{bmatrix}
\label{e2} 
\end{equation}

subs (\ref{e2}) in to (\ref{e1}), we get
\begin{equation}
  \begin{aligned}
  \psi &=  atan2( t_{6}sin(q7-al),- r_{6}sin(q) -t_{6}cos(q)cos(q7 - al))+pi \\
       &=  atan2( t_{6}sin(q8),- r_{6}sin(q) -t_{6}cos(q)cos(q8))+pi 
  \end{aligned}
  \label{e3}
\end{equation}
where $q8 = q7-al$

\paragraph{pose equation}
denote the point S in frame $F_{6}$, $^{6}S=(S_{x},S_{y},S_{z},1)$
according to DH, 
\begin{equation}
   ^{3}S = col_{4}(^{3}T_{2}) = \begin{bmatrix}
0 \\
0 \\
d_{se} \\
1
\end{bmatrix}
\label{ep1}   
\end{equation}
Additionally, 
\begin{equation}
^{3}S  = ^{3}T_{4}*^{4}T_{5}*^{5}T_{6}*\ ^{6}S       
\label{ep2}
\end{equation}

combine (\ref{ep1}) and (\ref{ep2}) and square both side we get 
\begin{equation}
S_x^2 + S_y^2 + S_z^2 + 2d_{ew}S_{x}cos(q6) - 2d_{ew}S_{y}sin(q6) + d_{ew}^2   - d_{se}^2 = 0
\label{ep3}
\end{equation}

and we have $$ ^{0}S = (0, 0, d_{bs}, 1 ) $$ hence 
\begin{equation}
  ^{6}S =\ ^{6}T7*\ ^{0}T_{7}^{-1}*\ ^{0}S =
 \begin{bmatrix}
a_{wr}+d_{sc}sin(q)cos(q8) \\
d_{sc}cos(q) \\
d_{sc}sin(q)sin(q8) \\
1
\end{bmatrix}
\label{ep4}     
\end{equation}
subs (\ref{ep4}) in to (\ref{ep3}) we get:
\begin{equation}
   a_{wr}t_{6}-d_{sc}(r_{6}cos(q)- t_{6}cos(q8)sin(q)) - k =0   
\label{ep5}
\end{equation}
where \[ k = (a_{wr}^2 + d_{se}^2 - d_{sc}^2 - d_{ew}^2)/2\]
\paragraph{ quartic equation}
according the two constrains (\ref{e3}),(\ref{ep5}) we get two equations about q6 and q8 
we figure that these equations would reduced into a quartic equation about q6.
subs (\ref{ep5}) into (\ref{e3}) to reduce $cos(q8)$ get
\begin{equation}
      sin(q8) = \frac{(xcot(q)-d_{sc}r_{6}sin(q))*tan(psi)}{d_{sc}t_{6}}
      \label{ep6}
\end{equation}
where $x = a_{wr}*t_{6} - k - d_{sc}*r_{6}*cos(q)$\\
combine (\ref{ep5}),(\ref{ep6})with $sin(q8)^2 + cos(q8)^2 =1$
we get \[ x^2cos(psi)^2 + (xcos(q)sin(psi)-d_{sc}r_{6}sin(psi)sin(q)^2)^2 - (d_{sc}t_{6}cos(psi)sin(q))^2 = 0\]
expand the above equation and notice that $(t6 -a_{wr})^2 + r_{6}^2 = d_{ew}^2$
we get 
\begin{equation}
      tm1+ t_{6}tm2+ t_{6}^2tm3 = -r_{6}*y*(k-a_{wr}t_{6})
      \label{eq7}
\end{equation}
where \[ y = 2*d_{sc}cos(q) \]
\[ tm1 = cos(psi)^2 (k^2 - (a_{wr}^2 - d_{ew}^2) d_{sc}^2 cos(q)^2) + 
 1/2 (-2 a_{wr}^2 d_{sc}^2 + 2 d_{ew}^2 d_{sc}^2 + k^2 + k^2 cos(2q)) sin(psi)^2 \]
\[ tm2 = -2 a_{wr} cos(psi)^2 (k - d_{sc}^2 cos(q)^2) + 
  a_{wr} (2 d_{sc}^2 - k - k cos(2q)) sin(psi)^2\]
\[ tm3 = 1/8 (6 a_{wr}^2 - 8 d_{sc}^2 + 2 a_{wr}^2 cos(2 psi) - 
    a_{wr}^2 cos(2 (psi - q)) + 2 a_{wr}^2 cos(2 q) - 
    a_{wr}^2 cos(2 (psi + q)))\]

square (\ref{eq7}), we get
\begin{equation}
    g_{4}t_{6}^4 + g_{3}t_{6}^3 + g_{2}t_{6}^2 + g_{1}t_{6} + g_{0} = 0 
\label{eq8}
\end{equation}     
where \[ g_{4} = tm3^2 + a_{wr}^2 y^2 \]
\[ g_{3} = 2 tm2 tm3 - 2 a_{wr} (a_{wr}^2 + k) y^2 \]
\[ g_{2} = tm2^2 + 2 tm1 tm3 + (a_{wr}^4 - a_{wr}^2 (d_{ew}^2 - 4 k) + k^2) y^2 \]
\[ g_{1} = 2 tm1 tm2 - 2 a_{wr} k (a_{wr}^2 - d_{ew}^2 + k) y^2 \]
\[ g_{0} = tm1^2 + (a_{wr}^2 - d_{ew}^2) k^2 y^2 \]

(\ref{eq8}) is a quartic equation about t6, so using quartic solver we could 
analytically solve the equation and get four solution of q6, the (\ref{eq7}) could be used 
to get the sign of q6. 

subs q6 into (\ref{ep5}),(\ref{ep6}), we get sin(q8), cos(q8);
\[ q8 = atan2(sin(q8),cos(q8))\]
\[ q7 = q8 + al\]
\subsubsection{solve q4 q5}
we have 
\begin{equation}
  ^{6}S = col_{4}(^{6}T_{5}\ ^{5}T_{4}\ ^{4}T_{3}\ ^{3}T_{2}) \\
       = 
\begin{bmatrix}
 - d_{se}*(cos(q4)*cos(q6) + sin(q4)*sin(q5)*sin(q6)) - d_{ew}*cos(q6) \\
d_{se}*(cos(q4)*sin(q6) - cos(q6)*sin(q4)*sin(q5)) + d_{ew}*sin(q6) \\
d_{se}*cos(q5)*sin(q4) \\
1
\end{bmatrix}
\label{eq9}       
\end{equation}

square the two side we get 
\[ cos(q4) = (S_x^2 + S_y^2 + S_z^2 - d_{ew}^2 - d_{se}^2)/(2*d_{se}*d_{ew}) \]
combined (\ref{ep4}) then we get two solution of q4
\begin{equation}
      q4 = \pm acos((S_x^2 + S_y^2 + S_z^2 - d_{ew}^2 - d_{se}^2)/(2*d_{se}*d_{ew}))
\end{equation}

get the third rows of (\ref{eq9}) we get

we have 
\begin{align}
^{5}S &= col_{4}(^{5}T_{4}\ ^{4}T_{3}\ ^{3}T_{2})
       = 
\begin{bmatrix}
d_{se}*sin(q4)*sin(q5)\\
d_{se}*cos(q5)*sin(q4) \\
d_{ew} + d_{se}*cos(q4)\\
1
\end{bmatrix}
 \\
 &=\ ^{5}T_{6}*\ ^{6}S = 
 \begin{bmatrix}
 - S_{y}*cos(q6) - S_{x}*sin(q6)\\
 S_z \\
 S_{y}*sin(q6) - S_{x}*cos(q6)\\
1
\end{bmatrix}
\label{eq11}       
\end{align}

from the first two rows of (\ref{eq11})
we get 
\begin{equation}
      q5 = atan2((-S_{x}*sin(q6)-S_{y}*cos(q6)),S_{z})
\end{equation}

\subsubsection{solve q1 q2 q3}
first we have 
\begin{align}
^{0}R_{3}&=^{0}R_{1}(q_{1})*\ ^{1}R_{2}(q_{2})*\ ^{2}R_{3}(q_{3})\notag\\
&=\begin{bmatrix}
-c_{1}s_{2}s_{3}-c_{3}s_{1} & s_{1}s_{3}-c_{1}c_{3}s_{2} & -c_{1}c_{2}  \\
c_{1}c_{3}-s_{1}s_{2}s_{3} & -c_{3}s_{1}s_{2}-c_{1}s_{3} & -c_{2}s_{1} \\
-c_{2}s_{3} & -c_{2}c_{3} & s_{2} \\
\end{bmatrix}
\label{es1}
\end{align}

with $s_{i}=sin(q_{i}),c_{i}=cos(q_{i})$
Additionally, 
\begin{align}
^{0}R_{3}&=\ ^{0}R_{7}*\ ^{7}R_{3} \notag\\
&=^{0}R_{7}*\ ^{7}R_{6}(q_{7})*\ ^{6}R_{5}(q_{6})*\ ^{5}R_{4}(q_{5})*\ ^{4}R_{3}(q_{4})\notag\\
&\overset{\underset{\mathrm{denoted}}{}}{=}\{r_{ij}\}_{3*3}
\label{es2}
\end{align}

combine (\ref{es1}),(\ref{es2}) we get two solution of q2
\begin{equation}
q_{2}=\pm acos(r_{33})+\frac{\pi}{2}
\end{equation}
denote $sgn2 = sign(cos(q_{2}))$,
from $r_{13}$, $r_{23}$
\begin{equation}
q_{1}=atan2(-r_{23}*sgn2, -r_{13}*sgn2 )
\end{equation}
from $r_{31}$, $r_{32}$
\begin{equation}
q_{3}=atan2(-r_{31}*sgn2, -r_{32}*sgn2 )
\end{equation}

totally, 16(4*2*2) solutions would be get  
\subsection{general configuration}
The ik method above is only valid for a very small subset of all
possible end effector poses in the work space where the wrist
center C is located on the base z-axis and the 7th joint axis z7 is
located in the base xz-plane. However, a closer look on
the robot’s kinematics shows that all possible end effector
poses can be reduced to one of those special poses. \cite{tittel2023analysis}
shows an example of how general poses can be reduced to
a special pose which is similar to our case.  By freeing the first 3 joint (sphere joint). it would rotate
the robot around the sphere until it become a special configuration, it is easy to see that q4 q5 q6 q7 is 
state in such motion. so we could get the q4 to q7 with the same method, then using q4 to q7 to solve a sphere rotation to get q1 q2 q3.
the key step is get the 3 dof $d_{sc}$, q and al from given endpose;

\subsubsection{calculate dsc, q, al}
given general end frame $^{0}T_{7} =  \begin{pmatrix}
^{0}R_{7} & ^{0}P_{7} \\
0 & 1 \\
\end{pmatrix}$
\[ SC =\ ^{0}P_{7} - (0,0,d_{bs})^T \]
\[ d_{sc} = norm(SC), zv = SC/d_{sc}, z7 = col_{3}(\ ^{0}R_{7})  \]
\[ q = -acos( zv \cdot z7 )\]
\[ yv = z7 \times zv, yv = yv/norm(yv) \]
\[ x72 = yv \times z7 \]
\[ x7 = col_{1}(\ ^{0}R_{7})  \]
\[ al = atan2(dot(x72*x7,z7), dot(x72,x7)); \]

with the same method as in special configuration we get all the joints value, notice that in (\ref{es2}), the $^{0}R_{7}$ comes from
the original given $^{0}T_{7}$
\section{Singularity analysis}
the singularity analysis is also a basic part in robot, especilly in redundency robot, there are no simple way to get all singularities.Here we use the reciprocity-based method\cite{nokleby2001reciprocity} and give the result of the singularity:
\begin{itemize}
   \item Kinematic Singularities
        \begin{itemize}
         \item $q4 = \{0, \pi\}$
         \item $q2 = \pm\pi/2\  \&\&\  q3 =\{0, \pi\}$
         \item $q2 = \pm\pi/2\  \&\&\  q6 =\pm acos(-awr/dew)$
         \item $q5 = \{0, \pi\}\  \&\&\  q6 =\pm acos(-awr/dew)$         
        \end{itemize}
   \item Algorithmic Singularities
       \begin{itemize}
         \item $q2 = \pm\pi/2$
         \item $q6 =\pm acos(-awr/dew)$
         \item (dew + awr*cos(q6))*sin(q4) - awr*cos(q4)*sin(q5)*sin(q6)=0
       \end{itemize}
 \end{itemize}
\section{  IMPLEMENTATION AND RESULTS}
One main reason for the development of an analytical
solution for the IKP of the redundent robot is the ability
to have a real-time safe deterministic implementation that
can be used for moving the robot in any of its provided
control modes. A disadvantage of numerical solvers for the
IKP of redundant robots is, for example, their use of loops,
which makes the number of calculations unpredictable if
no maximum of iterations is specified. If such a maximum
is specified, there is guarantee that the algorithm finds a
solution within a given error threshold.
With the proposed analytical approach, the user will definitely get a correct solution if one exists and the maximum
number of calculations is always predictable regardless of the
given target. Even if the target lies outside the workspace,
the algorithm gives a solution for the IKP that is based on
geometric calculations being reasonable and predictable.
Another drawback of numerical solvers is the need for a
sophisticated initial guess which should be already close to
the solution. If the user just gives an arbitrary configuration
like the zero posture as initial joint configuration, the joint
values often diverge quite fast and end up in some fanciful
regions.
The C++ implementation only uses Eigen as external
library, which is known and popular for its real-time safety.
The methods and functions used for the implementation
of the proposed approach do also have a fixed number of
calculations and do not violate any real-time requirements.\\
Furthermore, we believe that the methods presented in the paper can inspire analytical inverse solutions with arm angle for Franka robots.

\bibliography{bib}
\bibliographystyle{ieeetr}

\end{document}